%% file: 00_main.tex
\documentclass[journal]{IEEEtran} 
\usepackage{blindtext, graphicx}
\usepackage[framemethod=tikz]{mdframed}
\usepackage{amsfonts}
\usepackage{subfig}
\usepackage{tabularx}
\usepackage{color}
\usepackage{cite}
\usepackage[cmex10]{amsmath}
\usepackage{url}

\input{functions}

\begin{document}

\title{Data-Free/Data-Sparse Softmax Parameter \\ Estimation with Structured Class Geometries}
\author{\IEEEauthorblockN{Nisar Ahmed} \\
\IEEEauthorblockA{Ann and H.J. Smead Aerospace Engineering Sciences, \\ University of Colorado, Boulder, Colorado 80309\\
nisar.ahmed@colorado.edu
}
}

\maketitle

\begin{abstract}
This note considers softmax parameter estimation when little/no labeled training data is available, but a priori information about the relative geometry of class label log-odds boundaries is available. 
It is shown that `data-free' softmax model synthesis corresponds to solving a linear system of parameter equations, wherein desired dominant class log-odds boundaries are encoded via convex polytopes that decompose the input feature space. 
When solvable, the linear equations yield closed-form softmax parameter solution families using class boundary polytope specifications only. This allows softmax parameter learning to be implemented without expensive brute force data sampling and numerical optimization. 
The linear equations can also be adapted to constrained maximum likelihood estimation in data-sparse settings. 
Since solutions may also fail to exist for the linear parameter equations derived from certain polytope specifications, it is thus also shown that there exist probabilistic classification problems over $m$ convexly separable classes for which the log-odds boundaries cannot be learned using an $m$-class softmax model. 
\end{abstract}

\input{01_introduction}
\input{02_background}
\input{03_softmax_shaping}

\bibliographystyle{IEEEtran}
\bibliography{references}

\end{document}

%% file: functions.tex
\newcommand{\set}[1]{\left\{ #1 \right\}}

\newcommand{\esoftmaxfuncbias}[6]{\frac{e^{ #2_{#1}^T #3 + #6_{#1}}}{\sum_{#4=1}^{#5}{e^{ #2_{#4}^T #3 + #6_{#4}}}}} 
\newcommand{\RealSpace}[1]{\mathbb{R}^{#1}}

\newcommand{\pareqref}[1]{(\ref{eq:#1})}



%% file: 01_introduction.tex
\section{Introduction}\label{sec:introduction}
Softmax models are widely used to describe the `hybrid' conditional probability of discrete label-based outcomes given some continuous input/feature variable realization \cite{LernerThesis}. 
Such probability functions appear across many applications requiring differentiable `soft-switching' behaviors, such as: probabilistic and neural network classifiers \cite{williams1998bayesian, krishnapuram-pami2005-smlr, wang2018additivesoftmax}; stochastic hybrid dynamical system models \cite{Rui-SPLetters-2017, koller1999general}; semantic sensing models for soft data fusion \cite{Ahmed-TRO-2013}; constrained stochastic control and estimation \cite{brunskill2008continuous, wyffels2015negative}; and human decision models \cite{reverdy2016parameter, mcclelland2013probabilistic, Taguchi2009}. 

Yet, despite the softmax function's wide appeal, it remains surprisingly non-trivial in any domain to directly translate structured prior knowledge of hybrid mappings into softmax model parameters (let alone many other kinds of classifier model parameters, more generally \cite{Epshteyn-JAIR-2006}). 
Since softmax parameters are typically estimated from labeled training data via multinomial logistic regression \cite{Bishop06book}, one `brute force' approach is to simulate large synthetic auxiliary data sets which indirectly capture a priori class structure constraints or knowledge during training. 
However, this approach does not scale to large input feature spaces (e.g. where constraints may only apply in certain subspaces) or to likelihoods with dynamic class boundaries. 
Regression is also computationally expensive for large data sets, and prone to overfitting in data-sparse settings. Theoretically sound alternatives are therefore needed. 

Since softmax models are naturally useful for classification, their parametric model structures can be described geometrically by the probabilistic boundary sets between class labels in the input feature space. 
In particular, Theorem 3.1 of \cite{Taguchi2009} proved that, given any set of softmax parameters, the resulting log-odds boundaries between class labels always forms a full convex partition of the feature space. 
This partitioning leads to at most one convex log-odds polytope (CLOP) for each class label $i$, defining a convex input region where the class $i$'s probability dominates all other classes. 
This note sheds light on the above modeling issues by examining whether the reverse result holds: given an arbitrary convex decomposition of the input feature space describing a desired set of CLOPs (representing a priori structural information on the feature-based geometric relationships between class labels), is it always possible to find softmax parameters that embed it?  

It is shown that desired CLOPs can be directly embedded into softmax parameters whenever solutions to a corresponding system of linear equations exist. 
If solutions exist, then softmax parameters can be analytically identified or constrained using a priori structural knowledge (e.g. from set constraints that define forward hybrid dynamics \cite{asarin2010using, Blackmore-TRO-2010}, model space invariants, etc.). 
Expensive auxiliary data sampling and optimization can thus be avoided for data-free synthesis, 
and learning can be significantly improved in data-sparse cases.  
If solutions do not exist, then extended softmax models (i.e. multimodal softmax or hierarchical mixture of expert models) must be used to represent the underlying class boundaries in the same feature space (provided the CLOP specifications are realizable and irredundant). 
A related consequence of this result is that the true log-odds boundaries for certain convexly separable $m-$ary probabilistic classification problems are impossible to learn using an $m$ class softmax model, no matter how much training data is available. 
This could have interesting implications for other learning approaches that try to use a priori knowledge or invariant problem structures, e.g. \cite{hinton2015distilling, liu2016large}, as they do not account for the softmax model's limited representational power.  
Section II of this note describes technical preliminaries; Section III derives the main results, discusses their practical implications, and demonstrates these via application examples. 
Extending ideas initially presented in \cite{Sweet-ACC-2016}, this note provides significantly more technical depth and focus on the softmax synthesis problem and detailed illustrative examples for discussion. 

%% file: 02_background.tex
\section{Preliminaries}\label{sec:background}
Let $U \subseteq \RealSpace{n}$ be a continuous input feature space. 
Let $u \in U$ be an input vector realization, and let $D$ be a discrete random variable with possible categorical outcomes $\set{1,...,m}$, conditionally dependent on $u$. 
The hybrid conditional probability $P(D|u)$ can be modeled by the \emph{softmax function}, 
{
\allowdisplaybreaks
\abovedisplayskip = 2pt
\abovedisplayshortskip = 2pt
\belowdisplayskip = 1pt
\belowdisplayshortskip = 1pt
\begin{align}\label{eq:softmax}
P(D = i| u) = \esoftmaxfuncbias{i}{w}{u}{j}{m}{b} 
\end{align}
}
\noindent where $w_i \in \RealSpace{n}$ and $b_i \in \RealSpace{}$ are vector weight and scalar bias for class label $i$. Note that $u$ could be more generally replaced by some feature mapping $\phi(u)$; the scalar linear potential functions $a_i = w_i^Tu + b_i$ in the exponential arguments for each $i$ could also be generalized to arbitrary scalar non-linear potential functions of $u$ or $\phi(u)$ (as done in Gaussian process classification \cite{williams1998bayesian}). 
We refer to the input vector as $u$ without loss of generality, and consider only linear potential functions. 

Theorem 3.1 of ref. \cite{Taguchi2009} can be summarized as follows: for a given parameter set $\Theta = \set{w_c,b_c}_{c=1}^{m}$ of \pareqref{softmax}, the input space $U$ divides into $\tilde{m} \leq m$ convex `probabilistic polytopes', i.e. $\tilde{m}$ convex regions where the conditional probability for one class label dominates all others (such that each class has at most one dominant region). 
The linear boundaries of these polytopes at different relative probability levels are given by the set of log-odds functions between classes $i$ and $j$.
For $P(D=i|u) = P(D=j|u)$ (not necessarily equal to $0.5$), the log-odds boundary between $i$ and $j$ is
{
\allowdisplaybreaks
\abovedisplayskip = 2pt
\abovedisplayshortskip = 2pt
\belowdisplayskip = 1pt
\belowdisplayshortskip = 1pt
\begin{align}
\log \frac{P(D=j|u)}{P(D=i|u)} &= [\Delta w^T_{ji}, \Delta b_{ji}] \bar{u} =0,  \label{eq:logOdds}\\
\Delta w_{ji} = [w_j - w_i], \ \  & \Delta b_{ji} = (b_j - b_i), \ \ \bar{u} = [u,1]^T. \nonumber
\end{align}
}
\noindent This means the (non-unit) normal vector of the linear log-odds hyperplane between classes $i$ and $j$ is given by the \emph{difference} between the corresponding weight and bias parameters. 
Given $\Theta$, the convex log-odds polytope (CLOP) for class $j$ is defined by the set of $u$ for which the LHS of \pareqref{logOdds} is always positive with respect to all other labels $i \neq j$. 
 
Simple algebraic transformations to $\Theta$ can be used to shift, dilate, and rotate the log-odds boundaries within $U$. 
Eq. \pareqref{softmax} can also be extended to non-convex log-odds polytopes in several different ways. For instance, $u$ could be replaced with non-linear features $\phi(u)$. Non-convex log-odds polytopes can also be realized through a union of convex polytopes for a given class label, if softmax category labels can be grouped together as latent `subclasses' that uniquely map to observable classes of a \emph{multimodal softmax (MMS)} model \cite{Ahmed10a}. 
By using a sufficiently large number of subclasses, MMS models can approximate complex class boundaries via piecewise-convex log-odds polytopes. 
This idea is generalized further with hierarchical mixture of expert (HME) models \cite{Bishop06book}.  

Maximum likelihood or Bayesian methods can be used to estimate $\Theta$ from labeled training data \cite{Ahmed-TRO-2013}. 
The number of parameters scale linearly with the feature space dimension $n$. 
This makes the softmax function (and its extensions) more attractive and computationally tractable than alternative hybrid likelihood models such as unnormalized Gaussian mixtures \cite{brunskill2008continuous, wyffels2015negative}, which require many model terms to approximate large volume probability regions in high-dimensional spaces and have $O(n^2)$ parametric scaling per model term. 
However, as detailed in Sec. I, data-driven estimation is not ideal for incorporating a priori knowledge about class boundary properties. 
Furthermore, while optimization of likelihoods based on \pareqref{softmax} is convex \cite{mcfadden1974conditional}, MMS/HME model likelihoods are non-convex, which makes structured $\Theta$ estimation more challenging. 

%% file: 03_softmax_shaping.tex
\section{Data-free Softmax Parameter Identification}\label{sec:softmax_shaping}
Consider the `data-free softmax synthesis problem': given a set of $m$ discrete class (or subclass) labels with 1-to-1 correspondence to $m$ desired CLOPs in a complete mutually exclusive convex decomposition of input feature space $U = \RealSpace{n}$, find the model parameters $\Theta$ that produce log-odds boundaries for the desired CLOPs. 
The synthesis problem is stated here in its most restrictive form, i.e. with the strong assumption that the desired CLOPs for all $m$ class labels are available without any labeled training data. 
This is a reasonable starting point in light of Theorem 3.1 of \cite{Taguchi2009}, as it provides valuable insights for weaker forms of the synthesis problem, in which the CLOPs for class labels are only partially specified and/or limited training data is available. 

Let $ {\cal B} = \set{n^T_{ji}, c_{ji}}$, where, for $n_{ji} \in \RealSpace{n}$,  $c_{ji} \in \RealSpace{}$, and $i,j \in \set{1,...,m}$, the relations 
{
\allowdisplaybreaks
\abovedisplayskip = 2pt
\abovedisplayshortskip = 2pt
\belowdisplayskip = 1pt
\belowdisplayshortskip = 1pt
\begin{align}
[n^T_{ji}, c_{ji}] \bar{u} =0 \label{eq:desired_normals}
\end{align}
}
\noindent define a set of surface normal parameters describing the desired probabilistic class boundary hyperplanes for the faces of $m$ non-overlapping convex polytopes, which taken together form a complete mutually exclusive convex decomposition of $U$. Note that $n_{ji}$ is not necessarily a normalized unit vector. 
Assume w.l.o.g. that these implied polytopes correspond to desired dominance regions for classes $i,j \in \set{1,...,m}$, such that the dominance region for $j$ is given by the set of $u$ for which the LHS of \pareqref{desired_normals} is positive for all $i$. 
Hence, the relations \pareqref{desired_normals} in ${\cal B}$ define the hyperplanes at which the desired dominance regions of $j$ and neighboring labels $i$ intersect (i.e. the desired equiprobability level sets). 
Let $S_{\cal B}$ be the sets of neighboring class labels pairs $(i,j)$ whose CLOP boundaries are in ${\cal B}$. 

The data-free softmax synthesis problem can then be re-stated as: find softmax parameters $\Theta = \set{w_c,b_c}_{c=1}^{m}$ such that the log-odds boundaries for $i,j \in \set{1,...,m}$ given by \pareqref{logOdds} correspond exactly to the specified boundaries in ${\cal B}$. Equating the LHS of \pareqref{desired_normals} and \pareqref{logOdds}, $\forall (i,j) \in S_{\cal B}$
{
\allowdisplaybreaks
\abovedisplayskip = 2pt
\abovedisplayshortskip = 2pt
\belowdisplayskip = 1pt
\belowdisplayshortskip = 1pt
\begin{align}
&[\Delta w^T_{ji}, \Delta b_{ji}] \bar{u}_k = [n^T_{ji}, c_{ji}] \bar{u}_k \nonumber \\
\Rightarrow \  & w_{j} - w_{i} = n_{ji}, \ \ b_{j} - b_{i} = c_{ji} 
\label{eq:diff_constraints}
\end{align}
}
If $S_{\cal B}$ defines $N_S$ boundary pairs, then eq. \pareqref{diff_constraints} leads to a system of $N_S(n+1)$ linear equations that must be satisfied by $\Theta$. Define the stacked parameter vector $\vec{\theta} = [w_{1}^T, b_{1}, w_{2}^T, b_{2},...,w^T_{m}, b_{m}]^T$, and the stacked vector of desired polytope boundary surface normal parameters $\vec{\beta} = [n_{v(1)}^T, c_{v(1)}, n_{v(2)}^T, c_{v(2)},...,n_{v(N_S)}^T, c_{v(N_S)}]^T,$ where $v(t)$ denotes the $t^{\mbox{\tiny th}}$ pair of $(i,j)$ class labels in $S_{\cal B}$ for $t \in \set{1,...,N_S}$. 
Then, the required system of equations is 
\begin{align}
\allowdisplaybreaks
\abovedisplayskip = 1pt
\abovedisplayshortskip = 1pt
\belowdisplayskip = 0pt
\belowdisplayshortskip = 0pt
M\vec{\theta} = \vec{\beta}, \label{eq:linsyseqs}
\end{align}
\noindent where each row of matrix $M \in \RealSpace{N_S(n+1) \times m(n+1)}$ performs the required differencing operations on $\vec{\theta} \in \RealSpace{m(n+1)}$ to obtain $\vec{\beta} \in \RealSpace{N_S(n+1)}$ (the non-zero entries of $M$ are only $1$ or $-1$). 
 
Hence, depending on the desired CLOP specifications $({\cal B}, S_{{\cal B}}) \rightarrow (\beta, M)$, the existence and uniqueness of a solution $\vec{\theta}$ to any instance of the data-free softmax synthesis problem is immediately deduced via the ranks $r(M)$ of $M$ and $r(A)$ of the augmented matrix $A = [M,\vec{\beta}]$: 
\begin{enumerate}	
	\item $r(A) \neq r(M) \Leftrightarrow$ no exact solutions; 
	\item $r(A) = r(M) = m(n+1) \Leftrightarrow$ unique solution;
	\item $r(A) = r(M) < m(n+1) \Leftrightarrow$ infinitely many solutions.
\end{enumerate}
Each case is discussed next in more detail, with examples to illustrate the structured synthesis approach for $U = \RealSpace{2}$.

\subsection{Case 1: Inconsistent CLOP specifications}
This case implies that certain sets of desired CLOPs for $m$ class labels cannot be embedded in $\vec{\theta}$. 
Therefore, the data-free softmax synthesis problem does \emph{not} always have a solution, and the converse of Theorem 3.1 of \cite{Taguchi2009} does not generally hold. But if any $\Theta$ produces a convex decomposition of $U$, why should certain convex decompositions not have any corresponding $\Theta$? One geometric interpretation is that the magnitudes and directions of the desired non-unit hyperplane normals $n_{ji}$ (i.e. the desired probability gradient between classes $j$ and $i$ via $\Delta w_{ji}$ and $\Delta b_{ji}$) must satisfy an internal balancing condition inherent to all `feasible' softmax models. 
This follows from dependencies between parameters for any pair $i$ and $j$, which arise even if the dominant regions for $i$ and $j$ are not neighbors according to $S_{\cal B}$. 
Specifically, for a solution $\vec{\theta}$ to exist, it is easy to show that each $[\Delta w_{ji}, \Delta b_{ji}]$ and specified $[n_{ji}, c_{ji}]$ must obey a `loop summation constraint',
{
\allowdisplaybreaks
\abovedisplayskip = 2pt
\abovedisplayshortskip = 2pt
\belowdisplayskip = 1pt
\belowdisplayshortskip = 1pt
\begin{align}
&\Delta w_{j y_1} + \Delta w_{y_1 y_2} + \Delta w_{y_2 y_3} + ... + \Delta w_{y_{h}, i} + \Delta w_{ij} = 0 \nonumber \\
&\Rightarrow \ n_{j y_1} + n_{y_1 y_2} + n_{y_2 y_3} + ...  + n_{y_{h}, i} + n_{ij} = 0 \label{eq:nKVL}
\end{align} 
}
where \pareqref{nKVL} follows from \pareqref{diff_constraints}, and the indices $y_r$ for $r = 1,...,h$ and $2<h \leq m$ denote any subset of class indices other than $i$ and $j$ (similarly for $\Delta b_{ji}$ and $c_{ji}$). 
This is analogous to Kirchoff's votage law, i.e. the sum of differences between class activation potentials ($a_j$ and $a_i$) encountered in any closed label-to-label `loop' traversed within the softmax model must equal zero. 
Note that the $\Delta w$ constraints are automatically satisfied given any arbitrary $\vec{\theta}$ parameters. 
However, if the desired boundaries in ${\cal B}$ do not satisfy \pareqref{nKVL} for any subset of class labels $\set{i,j,y_1,...,y_h}$, then \pareqref{linsyseqs} has no solution using only $m$ softmax classes. 
The model must then be augmented to provide enough degrees of freedom in $\vec{\theta}$, so that the desired CLOPs for $m$ labels can be embedded via MMS or HME. 
Alternatively, a `best fit' approximation for an $m$-class model could be found, e.g. using a least-squares estimate $\vec{\theta}_{LS} = (M^TM)^{-1}M^T\vec{\beta}$.   

An interesting consequence is that there exist learning problems for which an $m$-class softmax model cannot exactly represent the linear log-odds boundaries for convexly separable classes in $U$. 
Thus, if case 1 arises in a `best case' learning scenario where the true CLOPs are known for all $m$ classes, then it will also be impossible to unbiasedly estimate $P(D|U)$ via multinomial logistic regression in a weaker learning scenario given only sampled data from the same set of classes. 

\subsubsection*{Example: T-boundary} Consider the CLOP specification for $m=3$ classes in $U = X_1 \times X_2$, shown by the dashed boundaries in Fig. \ref{fig:tboundaryexample} (left), where the desired normal vectors (dashed arrows) are given by $n_{12} = [-50,0]^T, n_{23} = [0,50]^T, n_{31} = [0,-50]^T $. 
This specification is inconsistent: while the $X_2$ components for the desired normals sum to zero, the $X_1$ components do not. 
Fig. \ref{fig:tboundaryexample} (left) also shows the resulting log-odds polytopes (colored regions) and normal vectors $\Delta w_{12}, \Delta w_{23}, \Delta w_{23}$ (solid arrows) for an $m=3$ softmax model obtained via the least-squares `best fit' estimator $\theta_{LS}$. 
This shows that basic softmax models can sometimes fail to represent even very simple convex probabilistic partitions of $U$. 
In this example, at least two `latent subclasses' are needed, such that their union provides the desired class CLOPs via an MMS model. 
Fig. \ref{fig:tboundaryexample} (middle) shows an augmented specification $n_{12} = [-50,0]^T, n_{24} = [0,50]^T, n_{34} = [50,0]^T, n_{31} = [0,-50]^T$ along with the resulting $m=4$ softmax model polytopes, where the original label class 3 is split into two new subclass labels 3 and 4. 
Fig. \ref{fig:tboundaryexample} (right) shows that the union of subclasses yields the desired $m=3$ CLOP specification. 
  
\input{fig_tboundary}

\subsection{Case 2: Unique solution}

\input{fig_example1polysynth}

\input{fig_smaxrangeexample}

In this case, $M$ is invertible, so $\vec{\theta} = M^{-1}\vec{\beta}$. 
Revisiting \pareqref{softmax}, it is possible to choose a basis for defining $n_{ji}$ and $c_{ji}$ such that $w_i=0$ and $b_i=0$ can be defined for some class $i$, and such that $M = I$ is an $(m-1)(n+1) \times (m-1)(n+1)$ identity matrix for $N_S = m-1$ (once $i$'s parameters are removed from $\vec{\theta}$) -- so that $\vec{\theta} = \vec{\beta}$ for the remaining parameters. 

\subsubsection*{Example: Polygon interior/exterior} Suppose it is desired to construct a likelihood model for being either `inside' or `outside' an arbitrary irregular polygon region in $U = X \times Y$, such as the one shown in Fig. \ref{fig:convpolygonex} (a). Since the `outside' region is non-convex, an MMS model with at least 5 softmax subclasses is needed to make up the entire `outside' class, with one subclass per polygon face (labeled `top left', `top right', `bottom left', `bottom right', and `bottom'). A single separate label is responsible for the `inside' class. Fig. \ref{fig:convpolygonex} (a) shows the desired polytope face normals $n_{ji}$ (black arrows) for each face between an `outside' subclass ($j$) and `inside' ($i$) class polytope. 
This specification of $N_s=5$ boundaries for $m=6$ labels with $n=2$ leads to a system of $N_S(n+1)=15$ linear equations in $m(n+1)=18$ parameters. Now let $w_i=0$ and $b_i=0$ for $i=$`inside', which gives 15 unknown softmax parameters in 15 equations. Thus, $M=I$, and $\vec{\theta} = \vec{\beta}$, i.e. the parameters for the 5 `outside' subclasses come directly from the polygon edge specifications in ${\cal B}$.  
Fig. \ref{fig:convpolygonex}(b) shows a top view of the resulting softmax probability surfaces, where the log-odds polytope boundaries coincide with the polygon edges, as desired. 
Fig. \ref{fig:convpolygonex}(c) shows the same softmax subclass surfaces from the side when each $n_{ji}$ has unit magnitude; Fig. \ref{fig:convpolygonex}(d) shows that the surface gradients become steeper when the magnitudes of each $n_{ji}$ are scaled up by a factor of 80, while the log-odds polytope boundaries remain the same. 
Fig. \ref{fig:convpolygonex}(e) shows the MMS union of subclass probability surfaces. 

\subsection{Case 3: Undetermined CLOP specifications}
In this case, there is only enough information available to partially constrain $\Theta$. 
If $\vec{\tau}$ is in the right null space of $M$, the family of solutions is given by $\vec{\theta} = M^T(MM^T)^{-1}\vec{\beta} + \vec{\tau}$, e.g. where $\vec{\tau}=0$ gives a minimum norm estimate. 
This case also allows equality constraints of the form \pareqref{linsyseqs} to be placed on $\Theta$ for numerical maximum likelihood or maximum a posteriori estimation with labeled training data. This provides an efficient principled way to embed a priori class structure information to constrain learning in data-sparse settings, without the need to generate and process additional synthetic training data. 

\subsubsection*{Example: Sparse semantic data modeling} Figure \ref{fig:likelihooddemo} shows how a priori information can guide calibration of semantic natural language data models in Bayesian soft sensor fusion for target localization \cite{Ahmed-TRO-2013}. 
Fig. \ref{fig:likelihooddemo} (a) and (b) show an `prototypical' model of a $m=17$ class semantic range and bearing model for describing the East-North location of a target relative to a fixed landmark (origin). 
These plots were generated via conventional softmax regression on synthetic training data, obtained by applying a rule-based labeling routine to a dense grid of sample points (the range limits for each class were arbitrarily selected for illustration only). 
An MMS `range-only' version of this semantic observation model with 3 classes is obtained by taking the union over semantic bearing labels in each of the range categories. 
Training data for calibrating the MMS model to human users in an actual target search experiment was obtained from sixteen human participants \cite{Ahmed-TRO-2013}; the markers in Fig. \ref{fig:likelihooddemo} (c) and (d) show a typical participant training data set, where only 32 data points could be obtained in a limited amount of time. 
Fig. \ref{fig:likelihooddemo} (c) shows the contours for the full 48-parameter MMS model learned directly from this data via numerical maximum likelihood optimization (using {\ttfamily fminunc} in Matlab). 
This approach clearly fails to capture the expected rotational symmetry and scale invariance of the underlying class regions (even with additional data provided from other human users to avoid the underspecification issue for parameter estimation). 
To embed this prior structure, eq. \pareqref{diff_constraints} can be used to define rotational invariance constraints on the log-odds boundary normals of the `nearby' and `far' classes. 
The resulting constraints of the form \pareqref{linsyseqs} effectively reduce $\Theta$ to 4 unique parameters, related to the steepness and displacements for the CLOPs of the `nearby' and `far' classes relative to the `next to' class.  
For instance, consider the inner `Nearby' ring of subclasses for `nearby N' (`nN'), `nearby NE' (`nNE'), etc. 
Suppose `nN' is designated as a template subclass with some set of nominal (but unknown) desired CLOP normal vectors and biases $\left\{n_{nN,NT},n_{nN,fN},n_{nN,nNE},n_{nN,nNW}\right\}$ (`NT': `Next To', `fN': `far North'). 
Rotational symmetry implies that, for the `nS' subclass, $n_{nS,NT} = {\cal R}(\pi) \cdot n_{nN,NT}, \ n_{nS,fN} = {\cal R}(\pi) \cdot n_{nN,fN}$, etc., where ${\cal R}(\theta)$ is the 2D rotation matrix about angle $\theta$. 
It follows that, if $n_{nN,NT}$, $n_{nN,fN}$, etc. were known, the $w_{nN}$ and $w_{nS}$ to be learned should satisfy $\Delta w_{nE,NT} = {\cal R}(\pi) \cdot \Delta w_{nN,NT} , \ \Delta w_{nE,fN} = {\cal R}(\pi) \cdot \Delta w_{nN,fN}$, etc. 
Extending this reasoning to the other subclasses mutatis mutandis gives a system of linear difference constraints in the form  
$M\vec{\theta} = \vec{\beta}$, where $\vec{\beta} = \Gamma \vec{\theta}$ and $\Gamma$ columns are the vectorized ${\cal R}(\theta)$ for template terms in $\vec{\theta}$. 
Fig. \ref{fig:likelihooddemo} (d) shows that the linearly constrained maximum likelihood estimate (using {\ttfamily fmincon}) provides much more satisfactory modeling results for the original sparse data set.

%% file: fig_tboundary.tex
\begin{figure}[t]
\centering
\includegraphics[width= 0.5\textwidth]{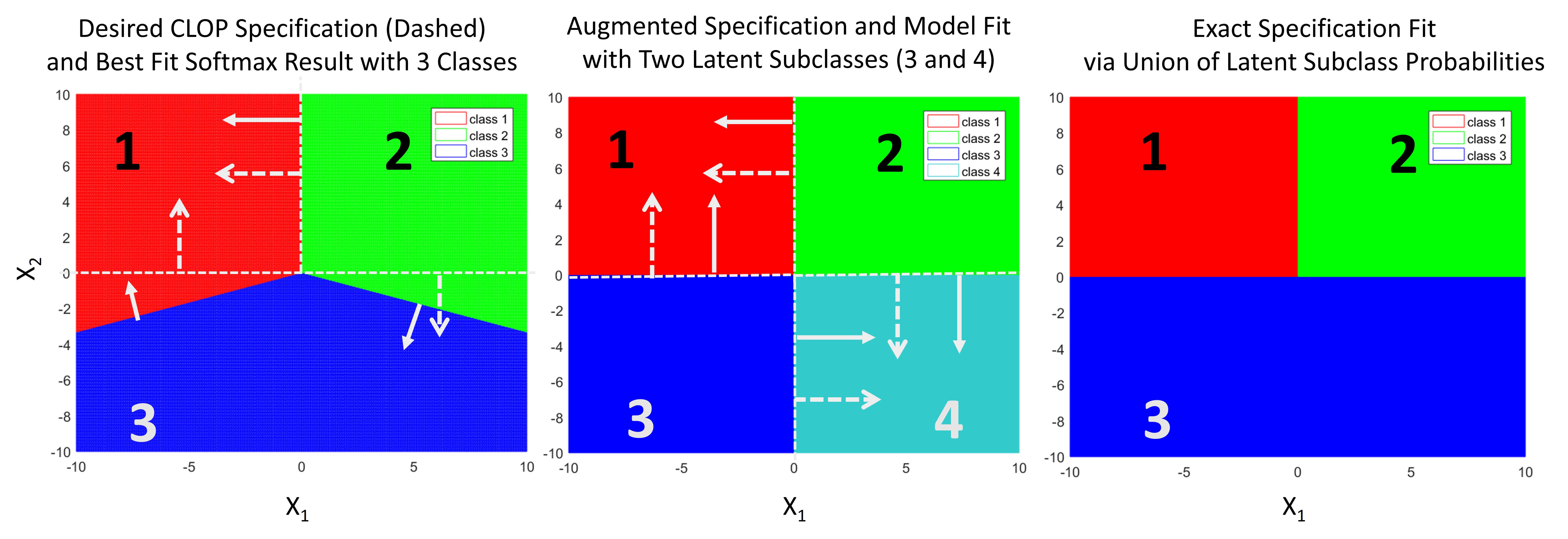} 
\caption{ \scriptsize T-boundary synthesis example (class probabilities nearly 1 in all regions).} 
\label{fig:tboundaryexample}
\vspace{-0.15 in}
\end{figure}

%% file: fig_example1polysynth.tex
\begin{figure*}[t!]
\centering
\begin{tabular}{@{}c@{}c@{}c@{}c@{}c@{}}
\includegraphics[width=4.05cm]{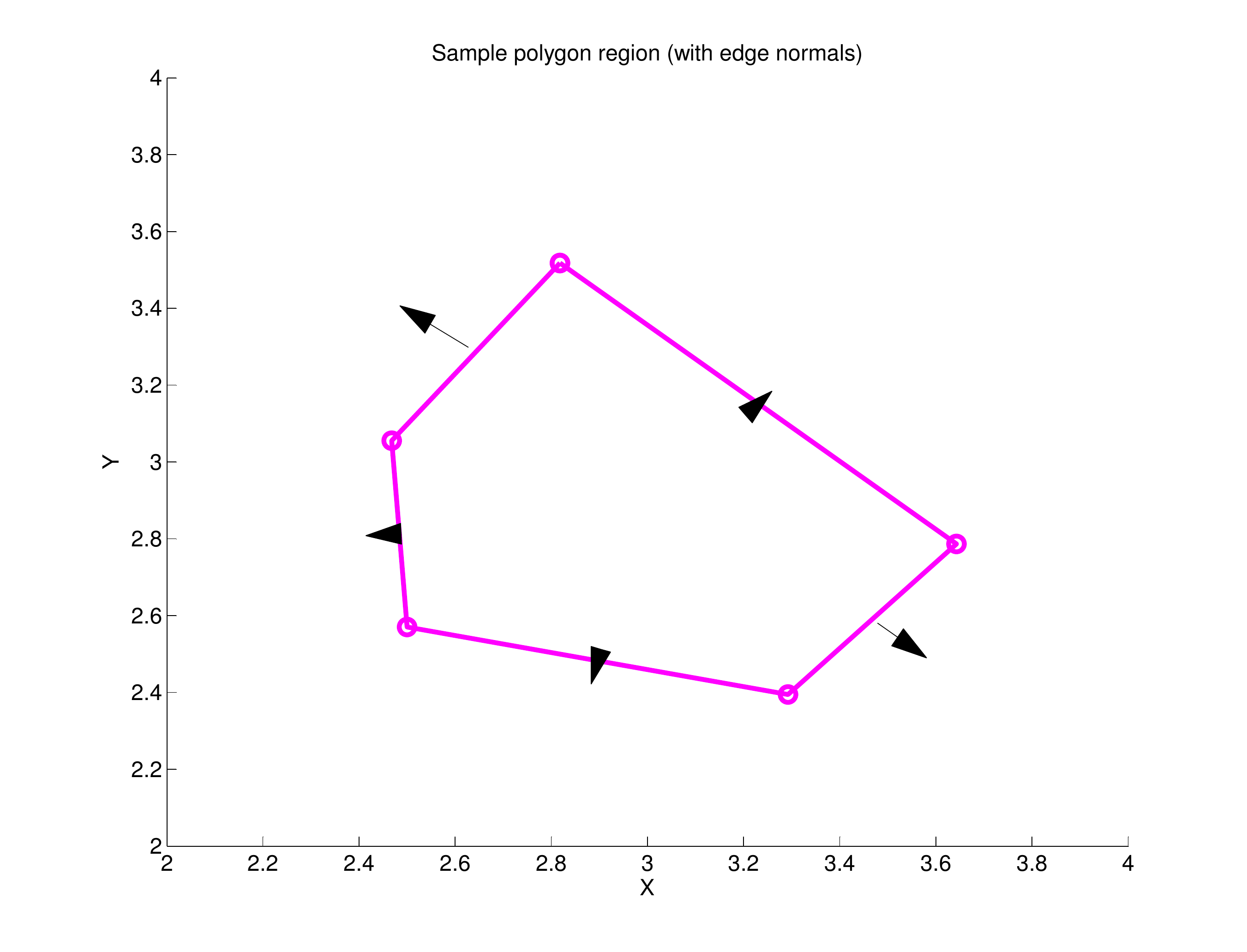}  &
\hspace{-0.5 cm}
\includegraphics[width=4.05cm]{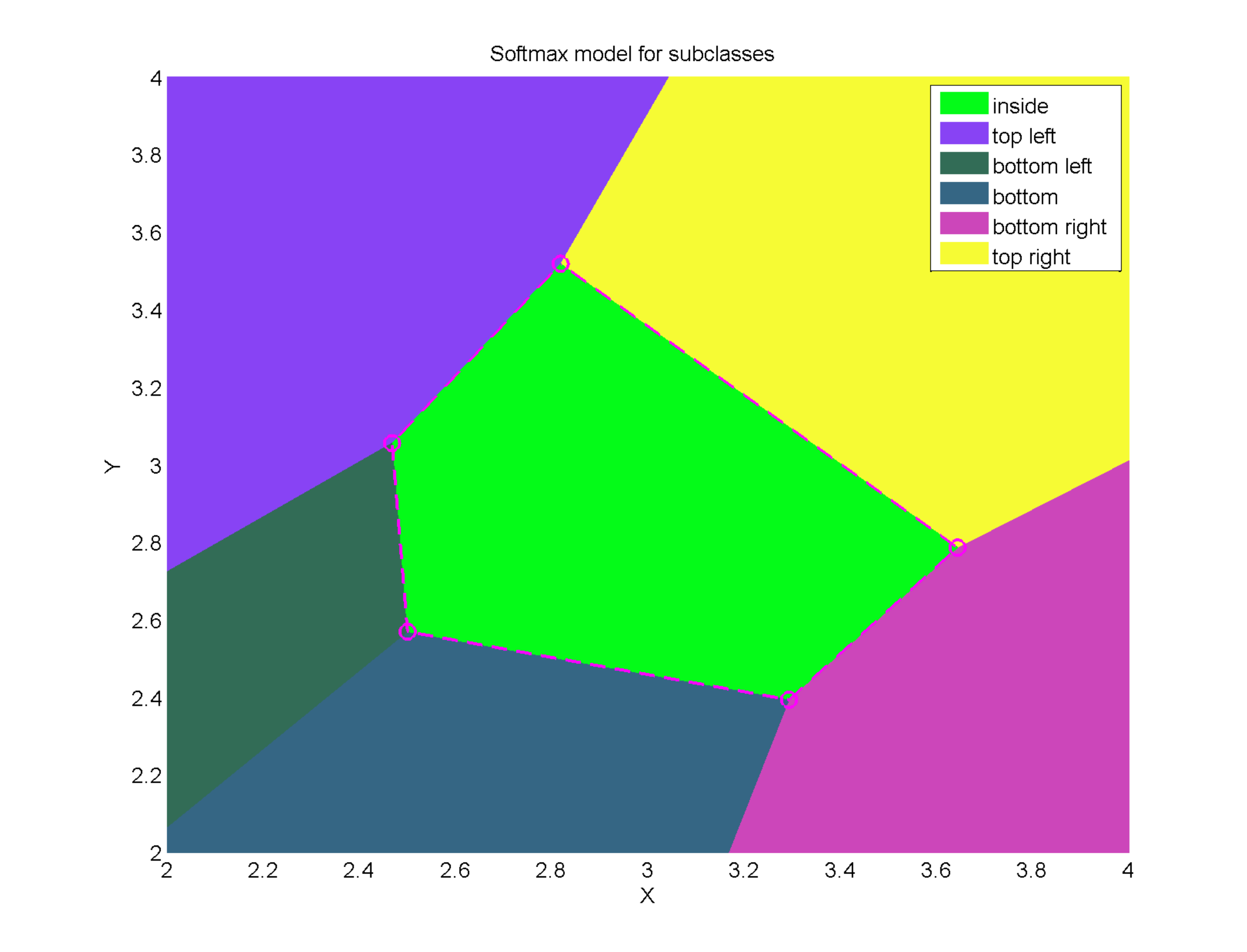}  &
\hspace{-0.5 cm}
\includegraphics[width=4.05cm]{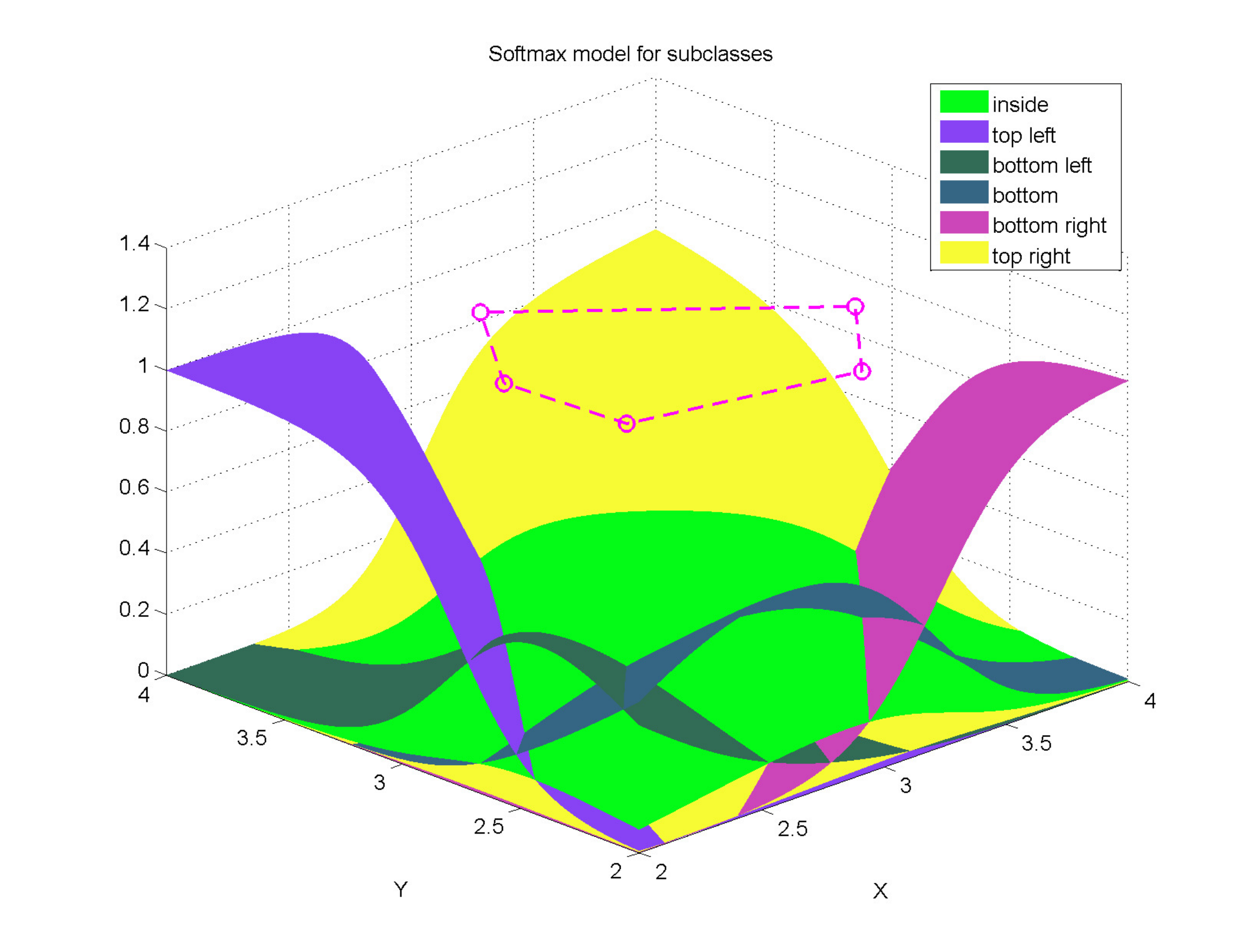}  &
\hspace{-0.5 cm}
\includegraphics[width=4.05cm]{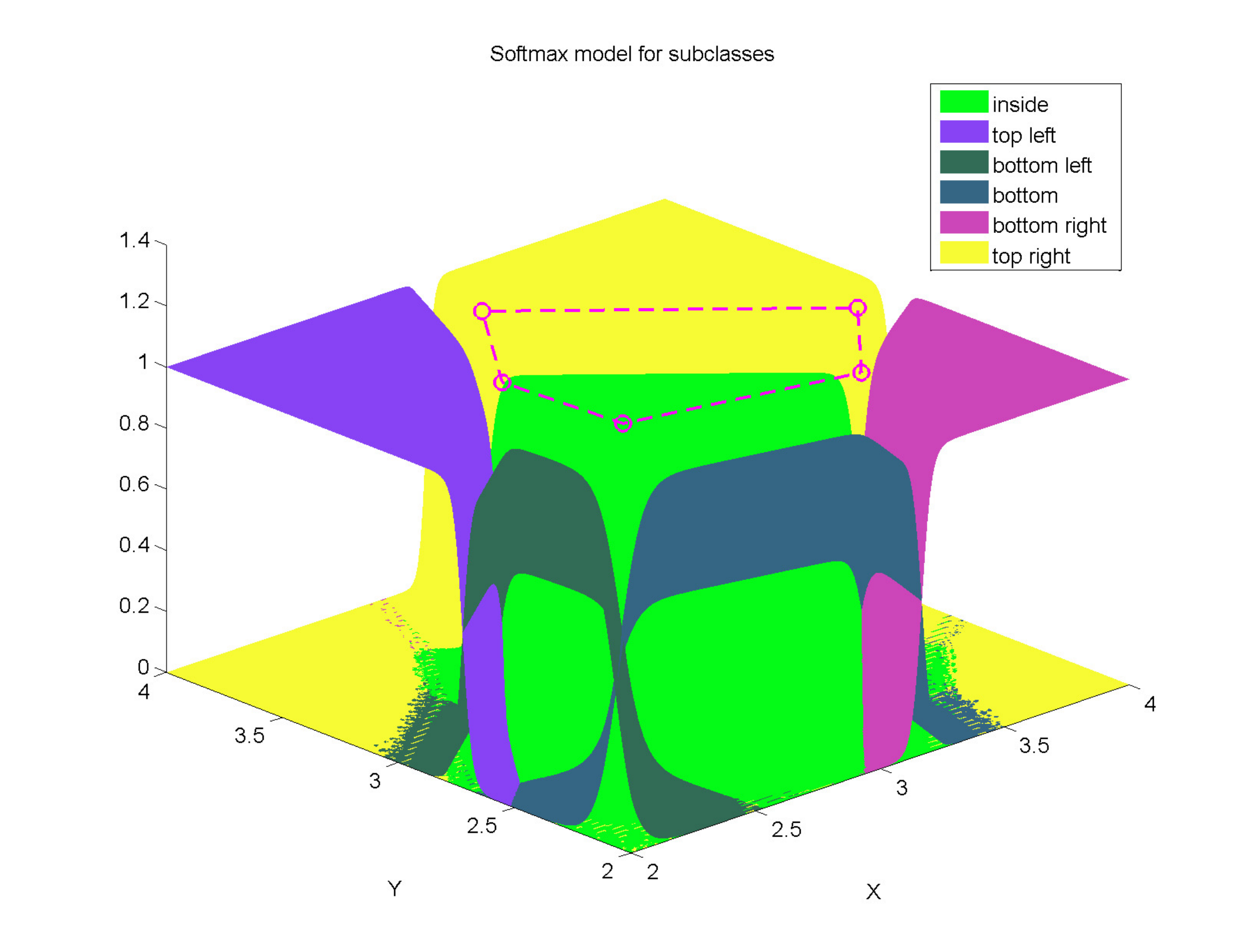} &
\hspace{-0.5 cm} 
\includegraphics[width=4.05cm]{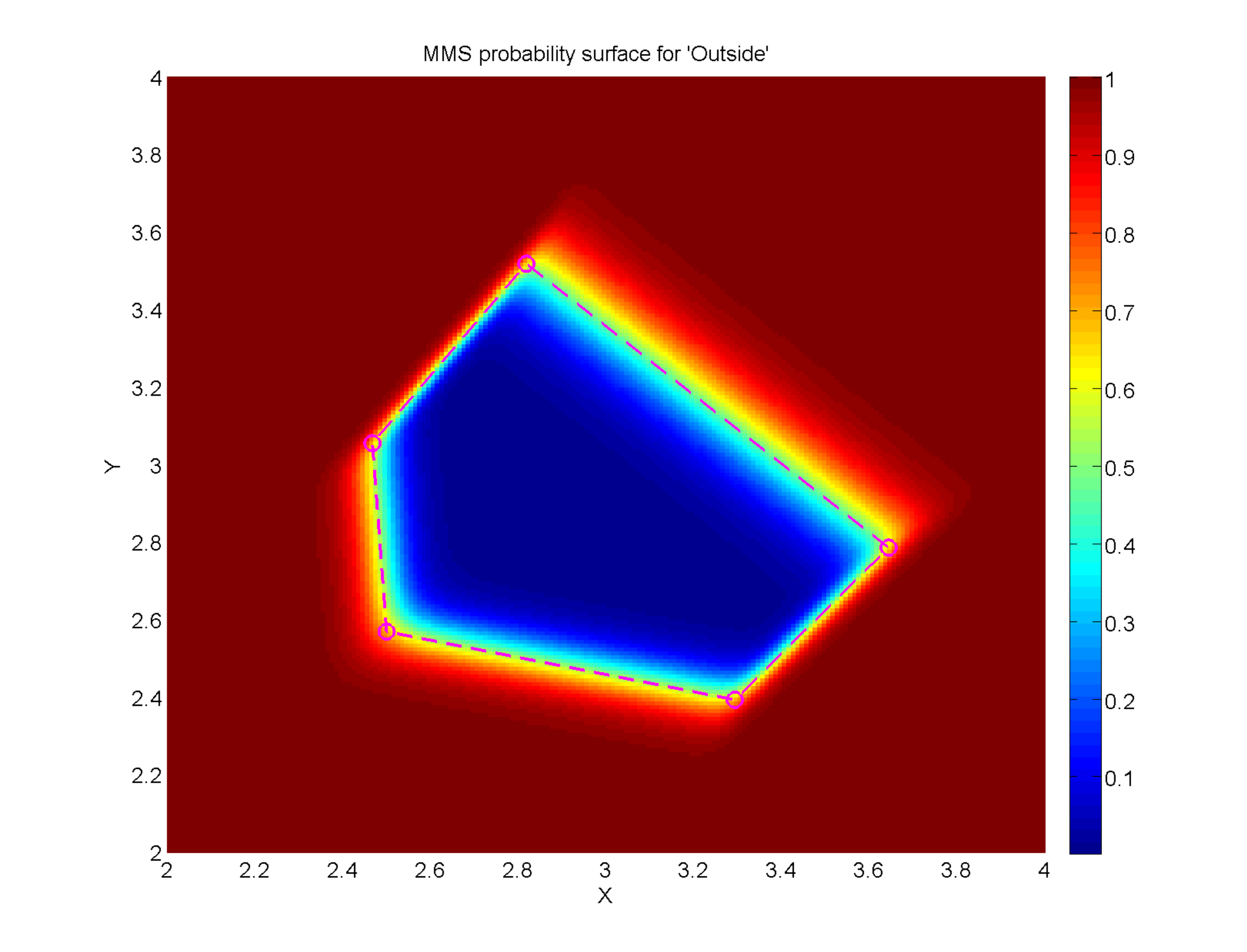} 
\\
\scriptsize (a) & \scriptsize (b) & \scriptsize (c) &\scriptsize (d) &\scriptsize (e)
\end{tabular}
\caption{\scriptsize (a) Specification; (b) resulting subclass regions; (c) subclass probabilities with unit normals $n_{ji}$; (d) desired normals magnifed by 80; (e) MMS likelihood for `outside'.}
\label{fig:convpolygonex}
\vspace{-0.1 in}
\end{figure*}

%% file: fig_smaxrangeexample.tex
\begin{figure*}[h!]
\centering
\begin{tabular}{@{}c@{}c@{}c@{}c@{}}
\includegraphics[width=4.95cm]{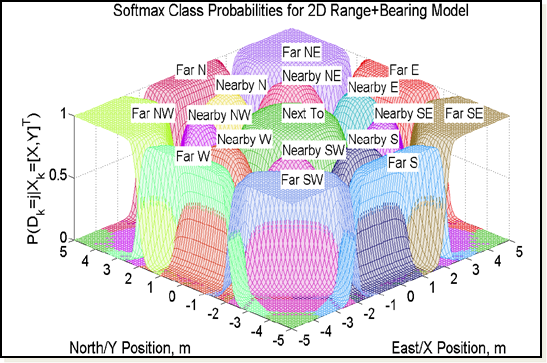} & 
\hspace{0.5cm}
\includegraphics[width=4.05cm]{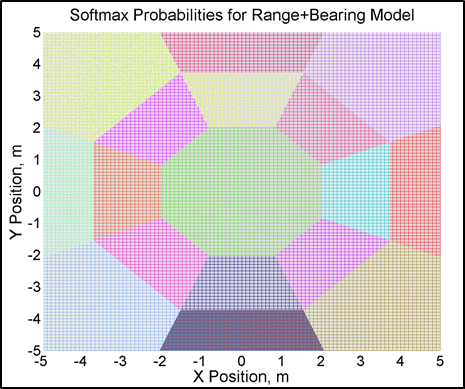}  &
\hspace{-0.5 cm}
\includegraphics[width=4.95cm]{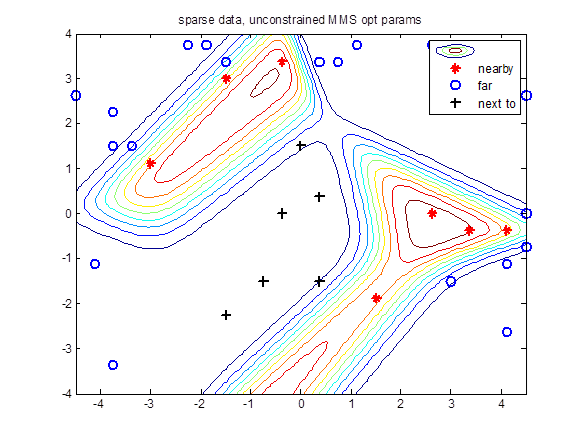}  &
\hspace{-0.5 cm}
\includegraphics[width=4.95cm]{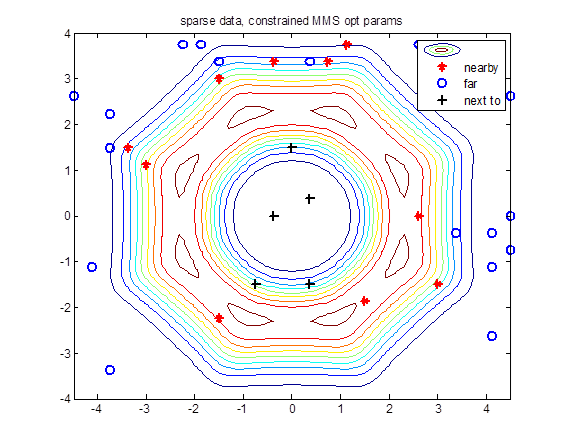}  \\
\scriptsize (a) & \scriptsize (b) & \scriptsize (c) &\scriptsize (d)
\end{tabular}
\caption{\scriptsize (a) Probability surfaces likelihood model where class labels take on a discrete range in {‘Next To’,‘Nearby’,‘Far From’} and
a canonical bearing {‘N’,‘NE’,‘E’,‘SE’,...,‘NW’}; (b) top down view of (a), showing convex log-odds class polytopes in $\mathbb{R}^2$; (c) probability contours of poor maximum likelihood learning results for 48 parameter range-only MMS model with 32 labeled training data points; (d) countours of improved maximum likelihood training results for constrained 4 parameter range-only MMS model with same data as in (c).}
\label{fig:likelihooddemo}
\vspace{-0.15 in}
\end{figure*}